\documentclass[runningheads]{llncs}
\usepackage[T1]{fontenc}
\usepackage{amsfonts}
\usepackage{amsmath}
\usepackage{booktabs}
\usepackage{url}

\usepackage{graphicx}
\usepackage{algorithm}
\usepackage{algorithmic}
\usepackage{multirow}
\usepackage{pifont} % for check and cross symbols
\usepackage{array}  % for better table formatting
\newcommand{\cmark}{\ding{51}} % check mark
\newcommand{\xmark}{\ding{55}} % cross mark
\usepackage{float}
\begin{document}
\title{Event-Frame Fusion for Inter-Frame Segmentation via Event-Guided Motion }%\\
%\vspace{1cm}{\normalfont Supplementary Material}}
\author{Dalia Hareb\inst{1,2}\orcidID{0009-0004-3022-8521} \and 
Jean Martinet\inst{1}\orcidID{0000-0001-8821-5556} \and
Benoit Miramond\inst{2}\orcidID{0000-0002-1229-7046} \and
Elisabetta Chicca\inst{3}\orcidID{0000-0002-5518-8990}}
%\author{}
\authorrunning{D.~Hareb et al.}
%\institute{}
\institute{I3S, Côte d’Azur University, CNRS, France \and
LEAT, Côte d'Azur University, France \\
\email{\{firstname.secondname\}@univ-cotedazur.fr} \and 
Zernike Institute, Groningen University, Netherlands\\
\email{e.chicca@rug.nl}}
\maketitle              % typeset the header of the contribution

\begin{abstract}
Autonomous navigation requires precise and efficient semantic segmentation, yet existing frame-based approaches remain limited by motion blur, glare, latency, and the low temporal resolution (20–30 FPS) of conventional cameras, which leads to information loss between frames. Event cameras have emerged as an alternative sensing modality, capturing intensity changes asynchronously with high temporal resolution, high dynamic range, and sparse outputs. However, event-based algorithms still fall short of frame-based ones in accuracy, as most segmentation methods are designed for dense frame data. To overcome these limitations, we propose a hybrid vision architecture that combines conventional frame-based and event-based cameras. The system integrates two complementary components: (1) a compact Spiking Neural Network (SNN) with 42k parameters for motion estimation, and (2) a lightweight event-driven SNN with 0.84M parameters for frame-based semantic segmentation, which interpolates motion between frames to refine segmentation results. By predicting inter-frame segmentations, the framework achieves segmentation rates of up to 500 Hz with an energy consumption below 1.87 mJ per inference, while maintaining real-time GPU execution at frequencies up to 200 Hz. Additionally, our approach compensates for information loss in frames affected by blur or overexposure, enabling more robust perception in challenging conditions.

\end{abstract}
% %
%
%
 
\section{Introduction} \label{intro}
Semantic segmentation enables fine-grained scene understanding, which is essential for autonomous navigation. Conventional frame-based methods, however, face several limitations. First, latency and temporal resolution are constrained by the camera frame rate, causing systems to miss critical dynamics occurring between consecutive frames. Simply increasing the frame rate could mitigate this problem but would also amplify bandwidth requirements \cite{gehrig_low-latency_2024}, computational complexity, and power consumption due to the redundancy inherent in consecutive frames. Second, motion blur and limited dynamic range degrade segmentation accuracy, particularly in challenging real-world lighting conditions.

\noindent These limitations are particularly relevant for advanced driver assistance systems using conventional cameras at 20 FPS, leading to blind times of 50 ms, which extend to 500 ms under adverse weather conditions. These gaps are critical in dynamic driving scenarios, where fast-moving objects must be detected and segmented in real time. For instance, during a 50–500 ms delay, a pedestrian running at 12 km/h and a cyclist at 20 km/h would move approximately 0.17–1.66 m, 0.27–2.77 m, respectively while a car at 50 km/h would travel 0.69-6.94 m during the same time interval, distances large enough to compromise safety-critical decisions. In contrast, performing segmentation every 2–10 ms (equivalent to 500–100 Hz) drastically reduces these blind distances to only 0.02–0.13 m, effectively saving 0.67–6.81 m of undetected motion. 
 
\noindent To overcome these challenges, we leverage event-based cameras, which capture per-pixel intensity changes with microsecond-level temporal precision, high dynamic range, low latency, and low power consumption \cite{gallego_event-based_2022}. While event-based algorithms alone often lag behind frame-based performance \cite{biswas_halsie_2023}, combining event-based sensing with conventional cameras enables both accurate and temporally dense semantic segmentation. Our approach introduces two lightweight spiking neural networks: a segmentation SNN (SegSNet) that produces frame-level predictions and a motion estimation SNN (SegMoSNet) that uses events to capture inter-frame object motion. The event-driven motion is then applied through a warping function to refine segmentation across time, producing temporally dense predictions without additional annotation or images. 

\noindent We adopt SNNs as our core computational paradigm due to their energy-efficient nature, offering a low-power alternative to conventional Artificial Neural Networks (ANNs), which are often computationally demanding \cite{dampfhoffer_leveraging_2023}, \cite{neftci_surrogate_2019}. These networks transmit information through spikes, generated only when a neuron's internal state (membrane potential) exceeds a threshold. Once this threshold is met, the potential resets to zero (hard reset) or is reduced by the threshold value (soft reset). Otherwise, it maintains its value (Integrate-and-Fire `IF' \cite{abbott_lapicques_1999}) or decays over time by a set factor (Leaky Integrate-and-Fire `LIF' \cite{abbott_lapicques_1999}). To train these networks, a differentiable approximation of the non-differentiable Heaviside activation function named Surrogate Gradient (SG) method \cite{neftci_surrogate_2019} is used. However, this approach can introduce estimation errors, leading to slower convergence compared to ANNs. 

\noindent To summarize, our contributions include a hybrid event–frame semantic segmentation framework that produces dense predictions at a frequency up to 500 Hz without additional annotations. The system relies on two compact SNNs (SegSNet: 0.84M parameters, 8.18 ms latency; SegMoSNet: 40k parameters, 1.18 ms latency) enabling real-time, energy-efficient inference ($<$1.5 mJ). An event-driven motion estimation network replacing optical flow to improve inter-frame segmentation consistency. Our approach also compensates for information loss in frames affected by motion blur or overexposure, supporting robust perception in challenging conditions.

\section{Related work}

Recent progress in neuromorphic semantic segmentation has focused on the use of spiking neural networks, event-based cameras, or hybrid approaches combining both. Early work by Alonso et al.~\cite{alonso_ev-segnet_2018} introduced Ev-SegNet, an encoder-decoder architecture based on Xception, tailored to process event streams directly. Gehrig et al. extended this by generating synthetic events from conventional video to improve training, as described in their paper~\cite{gehrig_video_2020}. Subsequent efforts have explored various enhancements for event-based segmentation. Wang et al.~\cite{wang_dual_2021} investigated event-to-image transformations to bridge the domain gap with frame-based methods. Sun et al.\cite{sun_ess_2022} proposed ESS, an unsupervised domain adaptation, while Wang et al.~\cite{wang_evdistill_2021}  and Xie et al.~\cite{xie_cross-modal_2024} introduced knowledge distillation techniques in their respective approaches, Evdistill and CMESS. In particular, Xie et al. used a cross-attention to softly align pseudo-pairs from the image and event domains. Hybrid models \cite{zhang_energy-efficient_2023} and HALSIE \cite{biswas_halsie_2023} that combine spiking and conventional artificial neural networks have also been developed. 
\\

\noindent In the fully spiking domain, Patel et al.~\cite{patel_spiking_2021} introduced a binary-class segmentation model using SNNs, which was later expanded by Kim et al.~\cite{kim_beyond_2021} to support multi-class segmentation using FCN and DeepLab, a time-aware encoder-decoder framework with Batch Normalization Through Time (BNTT). More recently, we proposed two complementary approaches: in \cite{hareb_evsegsnn_2024}, we introduced, EvSegSNN, a fully spiking architecture for event-based segmentation, while in \cite{hareb_enhanced_2025}, we improved the latency and energy efficiency of frame-based segmentation of our network by using events as a similarity metric. Finally, Zhang et al.~\cite{zhang_accurate_2024} introduced SpikeEDN, a spike-based neural architecture search method optimized for both raw event streams and fused event-frame inputs.
\\

\noindent The proposed approach differs from these works in three fundamental ways. Firstly, we directly exploit the intrinsic high temporal resolution of event data, in contrast to prior methods that accumulate all events between two frames to reconstruct a pseudo-frame, which is then processed similarly to standard RGB or grayscale images. Secondly, we have designed lightweight neural networks that are tailored to both frame-based and inter-frame semantic segmentation. This ensures that the networks are computationally efficient enough for real-time applications. Thirdly, we introduce motion segmentation as an alternative to optical flow, which is difficult to predict accurately and efficiently.

\section{Methodology}\label{method}
\subsection{Problem formulation}
Assume a sequence of $N+1$ RGB frames \(\{ I_0, I_1, \ldots, I_N \} \in \mathbb{R}^{H \times W \times 3}\) is captured by a 20Hz frame-based camera at timestamps \(\{ t_0, t_1, \ldots, t_N \}\), with corresponding per-frame ground-truth segmentation labels \(\{ L_0, L_1, \ldots, L_N \} \in \mathbb{R}^{H \times W \times C}\). Traditional frame-based approaches train a network to predict one segmentation map per frame \( I_i \). However, RGB frames suffer from information loss caused by motion blur or overexposure, decreasing the segmentation performances. To mitigate this issue, event-based cameras are integrated.
\\
Event cameras record asynchronous brightness changes at each pixel, outputting a stream of events. Each event \( e = [x, y, p, t] \) encodes the pixel coordinates \((x, y)\), the timestamp \(t\) at which the brightness change occurred, and the polarity \( p \in \{-1, 1\} \), indicating whether the brightness increased or decreased. 
\\
In existing approaches, events occurring between consecutive frame timestamps $[t_0, t_1], [t_1, t_2], \ldots, [t_{N-1}, t_N]$ are aggregated into event frames. These pseudo-frames are then fused with the corresponding RGB frames and processed for segmentation prediction. However, this integration still discards the fine-grained temporal data available between frames, resulting in a loss of information between frames.
\\
Our goal is to overcome these limitations by leveraging the high temporal resolution of event data.
By exploiting events captured over short intervals \(\Delta t\), we aim to generate inter-frame segmentations, effectively producing segmentation predictions at a frequency of \(\frac{1}{\Delta t}\) (e.g., $200Hz$ for $\Delta t=5ms$) without generating more than the available $N+1$ frames and ground-truth labels.

\begin{figure*}[t!]
    \centering
    \includegraphics[width=0.9\linewidth]{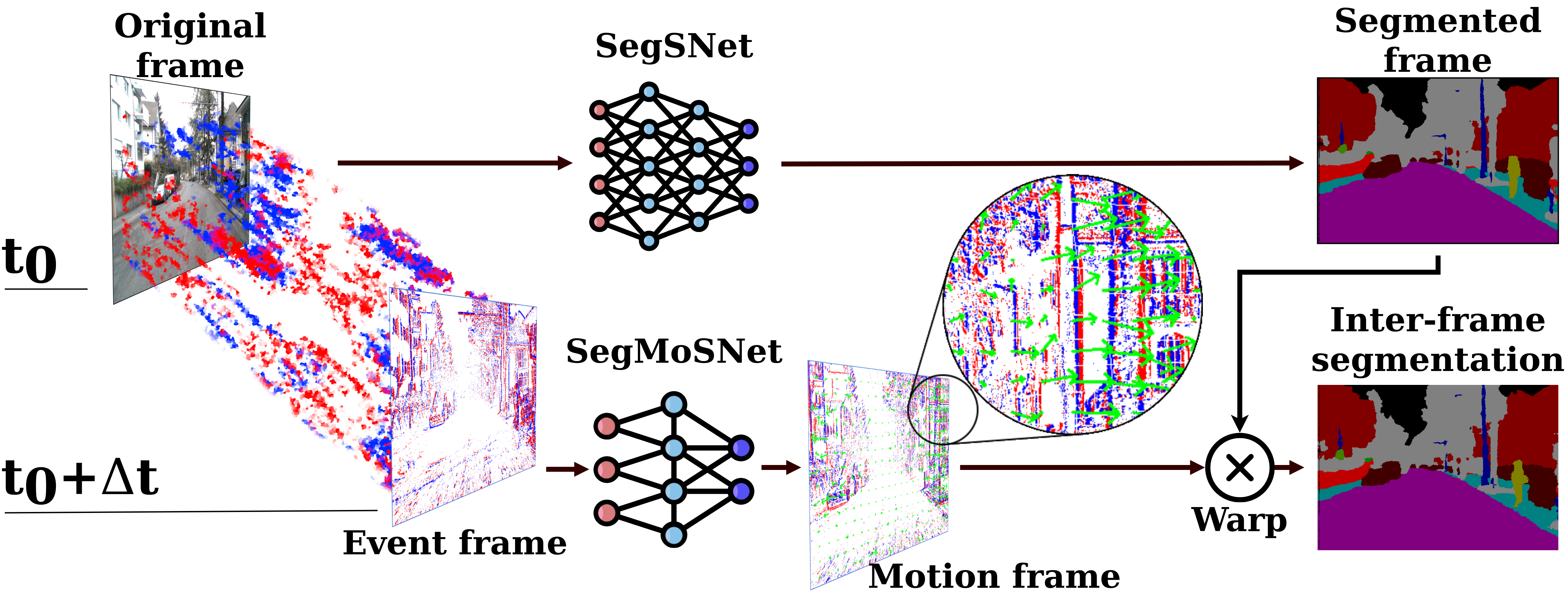}
    \caption {\textbf{Approach Overview:} the first frame at $t_0$ is segmented using SegSNet. Events accumulated over the interval $\Delta t$ are processed by the motion estimation network, SegMoSNet, to predict object motion during this period. The estimated motion, illustrated by green arrows on the zoomed-in motion frame, is then used to warp the previous segmentation and generate the updated segmentation at $t_0 + \Delta t$. }
    \label{overview}
\end{figure*}
\subsection{Method description } 
The proposed framework jointly exploits data from frame-based and event-based sensors, handled by two specialized spiking networks: SegSNet for semantic segmentation and SegMoSNet for motion estimation. This section details their respective architectures and explains how they are combined within a unified processing pipeline.

\begin{algorithm}[b]
    \begin{algorithmic}[1]
        \REQUIRE first frame: img, events stream $\in [t_i, t_{i+1}]$: events, $\Delta t$
        \ENSURE Inter-frame semantic segmentations
            \STATE seg $\leftarrow$ SegSNet(img)  
             \STATE seg $\leftarrow$ Upsample(seg)   
            \STATE acc\_events $\leftarrow $ Accumulate(events, $\Delta t$)
            \FOR{ ev in acc\_events}
                \STATE motion\_seg $\leftarrow$ SegMoSNet(ev)
                \STATE seg $\leftarrow$ warp(motion\_seg, seg)
                \STATE inter\_frame\_segs +=[seg]
            \ENDFOR
        \RETURN  inter\_frame\_segs
    \end{algorithmic}
     \caption{Inter-frame semantic segmentation.}
     \label{alg:event_seg}
\end{algorithm}

\subsubsection{Algorithmic approach details}
Our method, fully summarized in Algorithm \ref{alg:event_seg}, combines frame-based and event-based data to produce segmentations at every $\Delta t$ between two consecutive frames $I_i$ and $I_{i+1}$ captured at $t_i$ and $t_{i+1}$. The procedure, illustrated in Fig.\ref{overview}, begins by segmenting the initial frame $I_i$ using SegSNet generating the segmentation logits $S_{i,0}$. Subsequently, events triggered within $\Delta t$ after $t_i$ (i.e., $[t_i, t_i+\Delta t)$) are accumulated and fed into SegMoSNet to estimate motion. The resulting motion is then applied on $S_{i,0}$, after being upsampled using bilinear interpolation to match the original resolution, via a backward warping function \cite{zhu_deep_2017} and whose equation is defined in the supplementary material.
\\
This produces the first inter-frame segmentation logits $S_{i,1}$, as the warping step displaces pixel positions according to the observed motion within $\Delta t$. The segmentation mask, representing the per-pixel class, is generated by applying the argmax function on $S_{i,1}$
\\
The process is repeated for the next $\Delta t$ by: (1) accumulating events in $[t_i+\Delta t, t_i+2\Delta t]$ to estimate the motion during this interval and (2) applying the warping function to the first inter-frame segmentation logits $S_{i,1}$ using this newly estimated motion. This generates the second inter-frame segmentation logits $S_{i,2}$.
\\
This iterative procedure is executed for $T = \frac{t_{i+1} - t_i}{\Delta t}$ steps, progressively propagating and refining the segmentation across successive event intervals until $t_{i+1}$ is reached. Thus, the final segmentation $S_{i,T}$, obtained from events in $[t_i + (T-1)\Delta t, t_i+T\Delta t]$, is temporally aligned with the second frame $I_{i+1}$ at $t_{i+1}$ since $t_i + T\Delta t = t_{i+1}$. This temporal alignment represents the key idea for training SegMoSNet, as it enables the application of supervised learning.
\\
\noindent \textbf{Training procedure:} Once SegSNet has been trained on image frames using supervised learning according to Eq.\ref{loss}, its parameters are frozen. SegMoSNet is then trained on the event data by minimizing the loss between its final inter-frame segmentation and the ground-truth segmentation of the 2nd frame $I_{i+1}$, leveraging the established temporal alignment. Due to the recursive nature of our framework, where each intermediate segmentation is generated by warping the motion predicted by SegMoSNet on the previous segmentation, the model inherently forms a recurrent structure, illustrated in a figure available in the supplementary material. Therefore, backpropagating the error from the final inter-frame segmentation effectively performs backpropagation through time over a stateful spiking neural network unrolled across $T$ timesteps.
\\
\noindent \textbf{Evaluation procedure}: We further exploit temporal alignment by computing the Mean Intersection over Union (MIoU) between the final predicted segmentation and the ground-truth of the second frame. This measure reflects not only the accuracy of this segmentation but also the quality of the inter-frame segmentations, due to the cumulative nature of our approach. A high MIoU indicates accurate inter-frame segmentations, whereas a low MIoU suggests poor predictions. Direct evaluation of these intermediate segmentations is not possible, as their corresponding ground-truth annotations are unavailable.

\subsubsection{Network architecture description}
The segmentation network \textbf{SegSNet} processes dense RGB images using a spiking neural network with $0.84$ million parameters. Input frames are first transformed into binary spiking matrices via a $3\times3$ convolution (Conv), followed by batch normalization (BN) and spiking neurons (SNeuron). The resulting feature maps then pass through two additional Conv/BN/SNeuron layers and three residual Spike-Element-Wise blocks \cite{fang_deep_2022}. These blocks design enhances residual learning in deep SNNs while addressing the exploding and vanishing gradient problem. Finally, the feature maps pass through two more Conv/BN/SNeuron layers and reach the output layer.
This layer contains non-spiking read-out neurons. Each neuron simply integrates input spikes coming from the previous layer and loss function is calculated using real-valued potentials. The overall architecture is illustrated in supplementary material.

\noindent To address class imbalance and enhance segmentation accuracy, we adopt a composite loss function (Eq.\ref{loss}) that combines Dice loss \cite{milletari_v-net_2016} with the standard cross-entropy (CE) loss, both equally weighted. Dice loss refines the learning process by directly maximizing the overlap between predictions $\hat{y}$ and ground truth $y$, effectively optimizing the MIoU and improving the model’s ability to recognize minority classes.
\begin{equation}\label{loss}
    \begin{aligned}
       Loss &=  L_{CE} (y, \widehat{y}) + L_{Dice} (y, \widehat{y})  \\
         &= - \sum_{c=1}^{C} y_c\log(\widehat{y_c}) + 1 - \frac{1}{C} \sum_{c=1}^{C} \frac{2y_c\widehat{y_c} + 1}{y_c + \widehat{y_c} + 1}
    \end{aligned}
\end{equation}
\noindent The motion estimation network \textbf{SegMoSNet} illustrated in the supplementary material, follows a feedforward architecture with approximately $42K$ parameters. It consists of four sequential layers, each including a convolution, batch normalization and a spiking neuron. The final feature map is upsampled to the original image resolution using a deconvolution step involving 2D nearest-neighbor upsampling followed by a convolution. Similarly to SegSNet, the output layer contains non-spiking read-out neurons where the spikes coming from the previous layer are accumulated during $T$ timesteps. The network outputs a tensor matching the input resolution, with two channels encoding the displacement of each class along the x- and y-axes.
\section{Experiments}
\begin{figure}[t]
    \centering
    \includegraphics[width=0.49\linewidth]{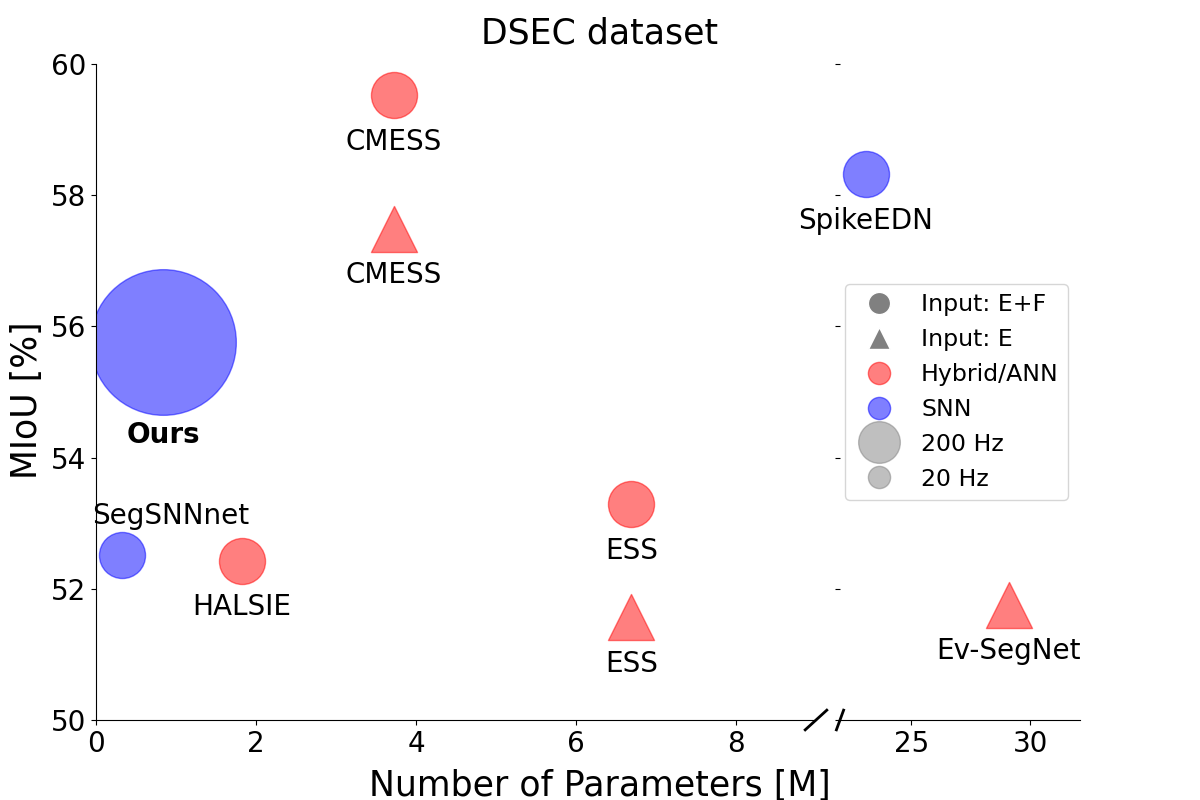}
    \hfill
    \includegraphics[width=0.49\linewidth]{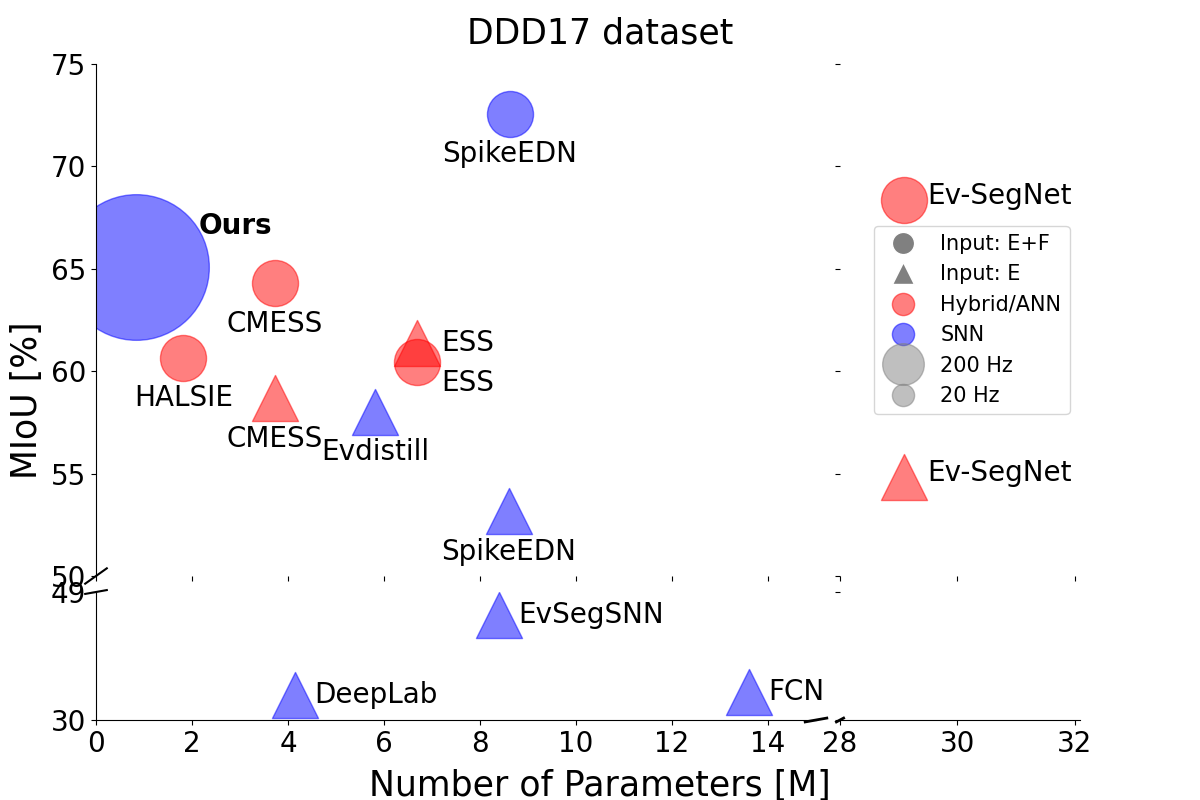}
    \caption{Evaluation of the proposed inter-frame segmentation approach on the DSEC-Semantic (left) and DDD17 (right) datasets, showing the best trade-off between segmentation accuracy, number of parameters and segmentation frequency.}
    \label{fig:sota_comparison}
\end{figure}
Evaluation is performed on two datasets, DSEC-Semantic and DDD17, where frames are generated every $50ms$. SegMoSNet is trained over $T=5$ timesteps, corresponding to $\Delta t = 10,ms$, while SegSNet is trained using a single timestep. Both networks utilize LIF neurons with hard reset. Gradients during backpropagation are estimated using a rectangular SG function \cite{wu_direct_2018}. All models are implemented in PyTorch and trained on an NVIDIA RTX A5000 GPU using the Adam optimizer, with a learning rate of $0.02$, for 100 epochs and batch sizes of 16 and 32 for SegMoSNet and SegSNet, respectively.

\subsection{Evaluation on DSEC-Semantic}
\begin{table}[!t]
\centering
\caption{Segmentation performance and event density (percentage of non-zero values in the histogram) across frequencies on DSEC-Semantic}
\begin{tabular}{l@{\hskip 6pt}c@{\hskip 6pt}c@{\hskip 6pt}c@{\hskip 6pt}c@{\hskip 6pt}}
\toprule
 &   \multicolumn{1}{c}{ w/o motion} & \multicolumn{3}{c}{w/ motion}  \\ \cmidrule(lr){2-2} \cmidrule(lr){3-5} 
 & 20Hz & 100Hz & 200Hz & 500Hz \\ \midrule
 % \multirow{2}{*}{TrainSet}  & 74.70\% & 74.21\%  & 73.03\%  & 68.72\%  \\
 %                            & (-0.00) & (-0.49) & (-1.37) & (-5.98) \\ \midrule
 \multirow{2}{*}{MIoU[\%]} & 56.28  & 56.12  & 55.76  & 53.35 \\
                              & (-0.00) & (-0.16) & (-0.52) & (-2.93) \\ \midrule
Density[\%]  & - & 8.88  & 5.41 &  2.67\\
\bottomrule
\end{tabular}

\label{tab:motion_freq}
\end{table}
\begin{table}[!t]
\centering
\caption{Impact of skipping frames on segmentation performance (MIoU) on the DSEC-Semantic dataset. Values in parentheses indicate the decrease in MIoU relative to no frame skipping.}
\begin{tabular}{l@{\hskip 6pt}c@{\hskip 6pt}c@{\hskip 6pt}c@{\hskip 6pt}c@{\hskip 6pt}}
\toprule
Skipped frames & 0 frame & 1 frame & 2 frames & 3 frames \\ \midrule
 % \multirow{2}{*}{TrainSet}  & 74.21\% & 72.11\%  & 71.41\%  & 71.08\%  \\
 %                            & (-0.00) & (-2.10) & (-2.80) & (-3.13) \\ \midrule
 \multirow{2}{*}{MIoU[\%]} & 56.12  & 56.71  & 55.19  & 54.79 \\
                              & (-0.00) & (+0.59) & (-0.93) & (-1.33) \\
\bottomrule
\end{tabular}

\label{tab:skip_frames}
\end{table}
The DSEC-semantic dataset, developed by Sun et al. \cite{sun_ess_2022} using Tao et al.'s hierarchical method \cite{tao_hierarchical_2020} consists of $8066$ training samples and $2803$ validation samples, captured across urban and rural environments in Switzerland with automotive-grade standard cameras and high-resolution event cameras. It includes $11$ classes:  background, building, fence, person, pole, road, sidewalk, vegetation, car, wall and traffic sign. Event streams are aggregated event-count image histograms \cite{maqueda_event-based_2018} per polarity, over a time window $\Delta t$, producing a tensor of shape $2 \times 640 \times 440$ after cropping the bottom 40 pixels, corresponding to the car’s dashboard.

\noindent\textbf{Results}: During inference, we vary the interval $\Delta t$ within $[2ms, 5ms, 10ms]$ to predict inter-frame segmentations at frequencies of $[500Hz, 200Hz, 100Hz]$. The results are summarized in Table \ref{tab:motion_freq}. The first row presents the MIoU scores, while the second row reports the average percentage of non-zero values per sample in the event histogram. The first column shows the performance of the standard approach, where segmentations are predicted every $50$ms using the segmentation network alone. After introducing the motion estimation network to produce higher-frequency segmentations, using the evaluation method detailed in Sec.\ref {method}, we observe that segmentation quality remains largely stable at $100$Hz and $200$Hz, but declines more noticeably at $500$Hz.

 \begin{figure}[tb]
    \centering
    \includegraphics[width=1.0\linewidth]{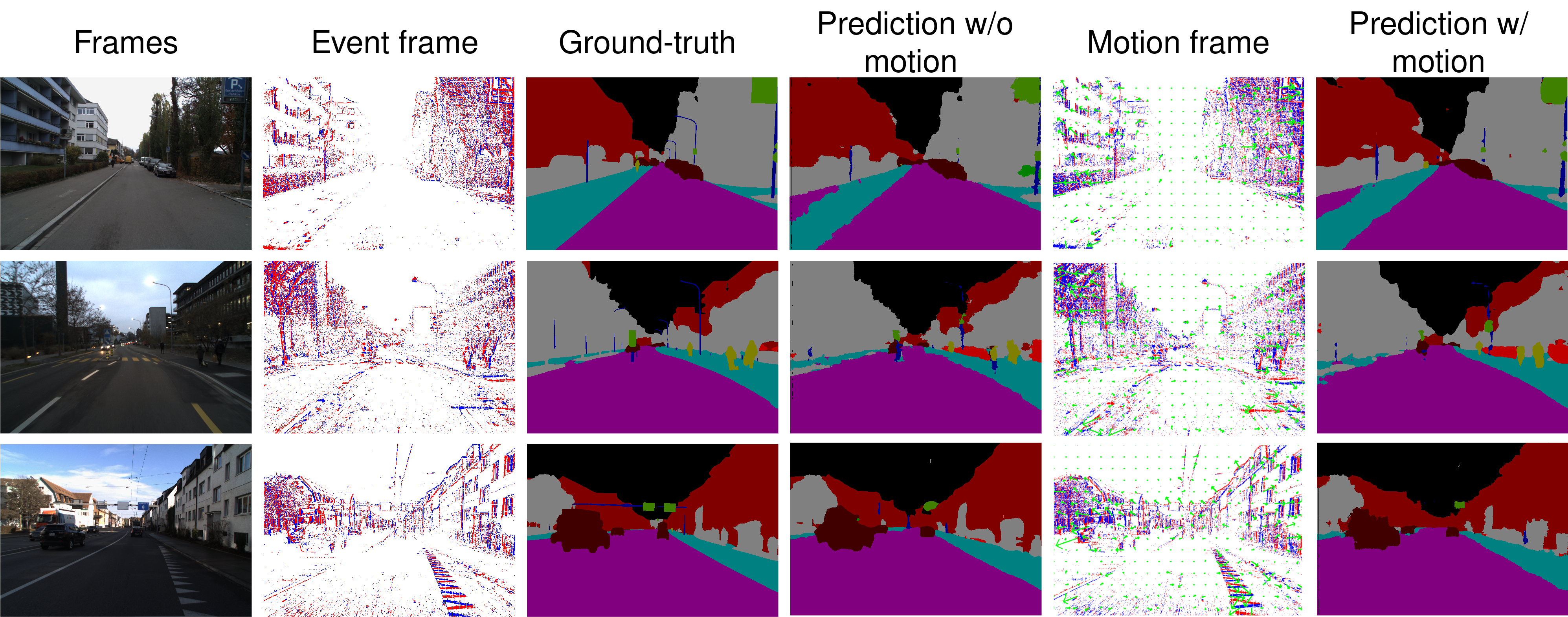}
    \caption{Qualitative results on the DSEC-Semantic dataset: From left to right, we display the following: the RGB image, accumulated events over the $10ms$ interval preceding the RGB capture, the ground-truth segmentation, the segmentation predicted using SegSNet alone, the estimated motion (visualized as green arrows overlaid on the event frame and computed from the $10ms$ events) and finally, the segmentation obtained using our motion-based approach.}
    \label{fig:qualitatif_dsec}
\end{figure}

This drop is likely attributed to the limited number of events generated within $2ms$, which accounts for only $2\%$ of the histogram, making reliable motion estimation more difficult. Qualitative results in Fig.\ref{fig:qualitatif_dsec} further highlight the minimal performance drop between segmentations predicted solely by SegSNet and those generated using the cumulative motion-based approach, confirming the high quality of inter-frame segmentation. 
The video\footnote{\url{https://youtu.be/zvs4-cEnLoI}} illustrates this process, displaying RGB frames at $20$Hz with ground-truth labels and SegSNet predictions on the top and inter-frame segmentations at $200$Hz along with the corresponding event streams on the bottom. illustrates this process, displaying RGB frames at $20$Hz with ground-truth labels and SegSNet predictions on the top and inter-frame segmentations at $200$Hz along with the corresponding event streams on the bottom.

\noindent In the previous experiment, we generated segmentations every $10ms$ for $100$Hz and periodically reset them every $50ms$ using SegSNet, which processes the corresponding frame captured at that moment. To further assess the robustness of our method, particularly in scenarios where frames suffer from motion blur or dazzling, we extended the evaluation by skipping $1$, $2$ and $3$ consecutive frames. In these cases, we continued the iterative pixel-shifting process until reaching the $3^\text{rd}$, $4^\text{th}$ and $5^\text{th}$ frames, corresponding to $100ms$, $150ms$ and $200ms$ after the initial reference frame. As shown in Table \ref{tab:skip_frames}, the MIoU loss increases slightly as more frames are skipped, with a maximum drop of only $1.33\%$ when 3 frames are skipped. This small degradation highlights the robustness of our approach, in addition to its capability for inter-frame segmentation.

\vspace{1em}  % add this between the two tables
\noindent It should be noted that motion estimation was not evaluated quantitatively due to the absence of ground-truth data. The only available reference is the optical flow provided in DSEC, computed over a $100$ms interval, which we use in the supplementary material to contrast with our estimated motion. To qualitatively illustrate the output, we present visualizations in Fig.\ref{fig:qualitatif_dsec}. Motion is represented by green arrows, scaled by a factor of $8$ and displayed with a stride of $20$ for clarity. In the video\footnote{\url{https://youtu.be/MkCc1_V009I}}, when the car turns left, the arrows point right, indicating the apparent motion of surrounding objects in the opposite direction. When the car moves straight, the arrows along the sides point backwards toward the image borders, while static regions (e.g., the sky) exhibit no motion, consistent with the absence of events in those areas.

\setlength{\tabcolsep}{3pt}
\begin{table}[t!]
\centering
\footnotesize
\caption{Comparison of our approach with the state-of-the-art methods using events (E) or events and frames (E+F) for semantic segmentation on DSEC-semantic}
\label{tab:methods_comparison}
\begin{tabular}{lcccccc}
\toprule
% \multirow{2}{*}{Method} & \multirow{2}{*}{Type} & \multirow{2}{*}{Input} & Params & MIoU & \multirow{2}{*}{FPS} \\
% &  &  & [M] & [\%] & \\ \midrule

Method&Type&Input&Params [M]&MIoU[\%]&FPS \\ \midrule
Ev-SegSNet \cite{alonso_ev-segnet_2018}  & ANN    & E     & 29.09 & 51.76 & 20  \\
ESS \cite{sun_ess_2022}       & ANN    & E     & 6.69  & 51.57 & 20  \\
ESS \cite{sun_ess_2022}       & ANN    & E+F   & 6.69  & 53.29 & 20  \\
CMESS \cite{xie_cross-modal_2024}      & ANN    & E     & 3.72  & 57.49 & 20  \\
CMESS \cite{xie_cross-modal_2024}    & ANN    & E+F   & 3.72  & 59.53 & 20  \\
 \midrule
HALSIE \cite{biswas_halsie_2023}   & Hybrid & E+F   & 1.82  & 52.43 & 20  \\ \midrule
SpikeEDN \cite{zhang_accurate_2024}   & SNN    & E     & 16.27 & 53.17 & 20  \\
SpikeEDN \cite{zhang_accurate_2024}   & SNN    & E+F   & 23.09 & 58.32 & 20  \\
SegSNNet \cite{hareb_enhanced_2025} & SNN    & E+F   & 0.33  & 52.52 & 20  \\
Ours       & SNN    & E+F   & 0.88  & 55.76 & 200 \\
\bottomrule
\end{tabular}
\end{table}

%illustrated in the top of Fig.\ref{fig:sota} and
\noindent \textbf{State-of-the-art comparison}: The left panel of Fig.~\ref{fig:sota_comparison} compares our approach with existing methods on the DSEC-Semantic dataset by plotting MIoU against the number of parameters. Segmentation frequency is indicated by the size of the points, the larger, the higher is the frequency. Red points represent ANN or hybrid networks, while blue points denote spiking neural networks. Triangles correspond to methods using only events (E) as input and disks indicate methods using both events and frames (F). A break on the x-axis is introduced to reduce large empty spaces between points. Detailed quantitative 
results are reported in Table~\ref{tab:methods_comparison}.
\\
Notably, our method is the only one that leverages the high temporal resolution of event data to produce segmentations at frequencies exceeding those of existing approaches, which typically operate at $20$Hz. Additionally, our approach achieves superior performance compared to most prior works while using fewer parameters, delivering high frame rates and being fully spiking. The only exceptions in MIoU performance are SpikeEDN \cite{zhang_accurate_2024}, which uses a 23M-parameter network that processes events and frames in parallel, with events converted into a continuous four-frame SBT \cite{wang_event-based_2019} spanning $200$ms and CMESS \cite{xie_cross-modal_2024}, which relies on a conventional ANN trained through a complex three-stage pipeline, including event-frame transformations and knowledge distillation and employs a Transformer network at inference.

\subsection{Evaluation on DDD17}

The DAVIS Driving Dataset (DDD17) for semantic segmentation is specifically tailored for automotive scenarios, encompassing 12 hours of driving data captured with a DAVIS camera. This dataset offers per-pixel aligned and temporally synchronized events alongside grey-scale frames. For ground-truth annotations, we rely on semantic pseudo-labels generated by Alonso et al. \cite{alonso_ev-segnet_2018}, obtained by pre-training the Xception network \cite{chollet_xception_2017} on frames extracted from the DAVIS dataset. The provided labels merge several classes, resulting in six distinct classes: flat (road and pavement), background (construction and sky), object, vegetation, human and vehicle. This dataset is divided into training and validation sets containing $15950$ and $3890$ grey-scale images, respectively, with dimensions of $260\times 346$, cropped to $200\times 346$. Similarly to DSEC dataset, event streams are converted into $2 \times 200 \times 346$ event-count image histograms over the interval $\Delta t$. 

\begin{table}[!t]
\centering
\caption{Segmentation performance and event histogram density across frequencies on DDD17.}
\begin{tabular}{l@{\hskip 6pt}c@{\hskip 6pt}c@{\hskip 6pt}c@{\hskip 6pt}c@{\hskip 6pt}}
\toprule
 &   \multicolumn{1}{c}{ w/o motion} & \multicolumn{3}{c}{w/ motion}  \\ \cmidrule(lr){2-2} \cmidrule(lr){3-5} 
 & 20Hz & 100Hz & 200Hz & 500Hz \\ \midrule

 \multirow{2}{*}{MIoU[\%]} & 67.05  &  66.75  & 65.08  & 59.88  \\
                              & (-0.00) & (-0.30) & (-1.97) & (-7.17) \\  \midrule
Density[\%]  & - & 1.97  & 1.07 &  0.4\\
\bottomrule
\end{tabular}
\label{tab:motion_freq_ddd17}
\end{table}
\begin{table}[!t]
\caption{Impact of skipping frames on segmentation performance (MIoU) on the DDD17 dataset. Values in parentheses indicate the decrease in MIoU relative to no frame skipping.}
\centering
\begin{tabular}{l@{\hskip 6pt}c@{\hskip 6pt}c@{\hskip 6pt}c@{\hskip 6pt}c@{\hskip 6pt}}
\toprule
Skipped frames & 0 frame & 1 frame & 2 frames & 3 frames \\ \midrule
\multirow{2}{*}{MIoU[\%]} & 66.75  & 62.98  & 61.38  & 59.77  \\
                              & (-0.00) & (-3.77) & (-5.37) & (-6.98) \\
\bottomrule
\end{tabular}
\label{tab:skip_frames_ddd17}
\end{table}

\noindent\textbf{Results:} The same set of experiments conducted on DSEC-Semantic was also performed on this dataset. Table \ref{tab:motion_freq_ddd17} shows that at $100$Hz, inter-frame segmentation retains good performance, with only a $0.30\%$ drop in MIoU. This loss increases to nearly $2\%$ at $200$Hz and rises sharply to $7.17\%$ at $500$Hz, likely due to the extremely low number of events within $2ms$ (only $0.4\%$), making motion estimation unreliable. 
To mitigate this issue, a promising direction for future work would be to make the approach dynamic by adapting $\Delta t$ based on event density.
Table \ref{tab:skip_frames_ddd17} illustrates that when the number of skipped frames increases, the MIoU decreases monotonically. This performance drop is due the poor quality of the DDD17 annotations that limits the effectiveness of long-range temporal propagation, as larger frame intervals increase the likelihood of inconsistencies between consecutive ground-truth labels.

%the bottom of Fig.\ref{fig:sota} and

\setlength{\tabcolsep}{3pt}
\begin{table}[!t]
\footnotesize
\centering
\caption{Comparison of our approach with the state-of-the-art methods on DDD17.}
\label{tab:methods_comparison2}
\begin{tabular}{lccccccc}
\toprule
% \multirow{2}{*}{Method} & \multirow{2}{*}{Type} & \multirow{2}{*}{Input} & Params & MIoU & \multirow{2}{*}{FPS} \\
% &  &  & [M] & [\%] & \\
% \midrule
Method&Type&Input&Params [M]&MIoU[\%]&FPS \\ \midrule
Ev-SegSNet \cite{alonso_ev-segnet_2018}   & ANN    & E     & 29.09 & 54.81  & 20 \\
Ev-SegSNet \cite{alonso_ev-segnet_2018}  & ANN    & E+F   & 29.09 & 68.36  & 20 \\
Evdistill \cite{wang_evdistill_2021}   & ANN    & E     & 5.81  & 58.02 & 20 \\
ESS \cite{sun_ess_2022}      & ANN    & E     & 6.69  & 61.37  & 20 \\
ESS \cite{sun_ess_2022}       & ANN    & E+F   & 6.69  & 60.43  & 20 \\
CMESS \cite{xie_cross-modal_2024}      & ANN    & E     & 3.72  & 58.69  & 20 \\
CMESS \cite{xie_cross-modal_2024}      & ANN    & E+F   & 3.72  & 64.30  & 20 \\
\midrule
HALSIE \cite{biswas_halsie_2023}    & Hybrid & E+F   & 1.82  & 60.66  & 20 \\ \midrule 
Spiking FCN \cite{kim_beyond_2021}      & SNN    & E     & 13.60 & 34.20  & 20 \\
Spiking DeepLab \cite{kim_beyond_2021}     & SNN    & E     & 4.14  & 33.70  & 20 \\ 
SpikeEDN \cite{zhang_accurate_2024}    & SNN    & E     & 8.60  & 53.15  & 20 \\
SpikeEDN \cite{zhang_accurate_2024}   & SNN    & E+F   & 8.62  & 72.57  & 20 \\
EvSegSNN \cite{hareb_evsegsnn_2024}    & SNN    & E     & 8.40  & 45.54  & 20 \\
Ours        & SNN    & E+F   & 0.88  & 65.08  & 200 \\
\bottomrule
\end{tabular}
\end{table}
\noindent\textbf{State-of-the-art comparison: }The comparison with state-of-the-art methods, reported in Table~\ref{tab:methods_comparison2} and illustrated in the right panel of Fig.~\ref{fig:sota_comparison}, highlights that our approach achieves the best trade-off between model size, segmentation accuracy and processing frequency. Specifically, it achieves a high segmentation frequency of $200$Hz with a compact model of only $0.88$M parameters, while outperforming most existing methods in terms of MIoU. Notable exceptions include EvSegSNet \cite{alonso_ev-segnet_2018} which employs a large 29.09M-parameter ANN processing events and frames in parallel and SpikeEDN \cite{zhang_accurate_2024} which, as explained in the DSEC-Semantic results description, uses a 8.62M-parameter network that similarly processes events and frames in parallel, with events converted into a continuous four-frame SBT \cite{wang_event-based_2019} over a $200$ms interval.

\subsection{Runtime and computational cost}

\begin{table}[!b]
\centering
\caption{Runtime comparison on GPU and CPU per inference. Real-time for an inter-frame segmentations predicted at up $200$Hz.}
\begin{tabular}{l@{\hskip 6pt}c@{\hskip 6pt}c@{\hskip 6pt}c@{\hskip 6pt}c@{\hskip 6pt}}
\toprule
 &SegSNet & SegMoSNet & Warping & Real-time  \\ \midrule
GPU &  8.18ms &  1.18ms  & 4.68ms & \cmark  \\
CPU & 34.22ms & 13.96ms & 31.93ms &\xmark \\                         
\bottomrule
\end{tabular}
\label{tab:runtime}
\end{table}

\noindent\textbf{Runtime}: We evaluated the runtime of SegSNet and SegMoSNet on both GPU and CPU by measuring the average inference time per prediction. As shown in Table \ref{tab:runtime}, SegMoSNet achieves significantly faster inference than SegSNet, thanks to its compact design with only $42$K parameters. On GPU, SegSNet performs full-frame segmentation in $8.18ms$, well within the $50ms$ interval between two consecutive frames, enabling real-time operation. For higher segmentation rates, such as $200$Hz (i.e., every $5ms$), SegMoSNet predicts motion from the first $5ms$ of events in just $1.18m$s. Once the initial segmentation is available, this motion is used to generate warped segmentations within $4.68ms$, following a brief $2ms$ delay. Subsequent event batches are processed immediately after being captured, as event collection runs in parallel with both motion prediction and warping. As a result, the system sustains real-time segmentation for frequencies up to $200$Hz. On CPU, runtime increases, primarily due to the cost of the warping operation. This overhead could be significantly reduced by exploiting the natural sparsity of event data: since motion is inherently localized in regions with events, predicting sparse motion maps would enable targeted warping on a small subset of pixels. As shown in Tabs.\ref{tab:motion_freq} and \ref{tab:motion_freq_ddd17}, events histograms are already highly sparse, exceeding $90\%$, indicating significant potential for computational savings.

% \begin{table}[b]
% \centering
% \footnotesize
% \begin{tabular}{l|@{\hskip 5pt}c@{\hskip 5pt}c@{\hskip 5pt}c@{\hskip 5pt}c@{\hskip 5pt}c@{\hskip 4pt}}
% \toprule
% \multirow{2}{*}{Dataset}& \multirow{2}{*}{FLOPs} & \multirow{2}{*}{Addr} & \multicolumn{3}{c}{Memory} \\ 
% \cmidrule(lr){4-6} 
%  &                      &                         & 8kb     & 32kb    & 1Mb     \\ \midrule 
% DSEC-Semantic & 0.32mJ               & 0.41mJ                & 110.19mJ &  220.39mJ & 1101.95mJ \\ \midrule
% DDD17 &  0.07mJ           & 0..09mJ               & 25.23mJ &  50.46mJ & 252.30mJ \\    
% \bottomrule
% \end{tabular}
% \caption{Computational cost of SegSNet and SegMoSNet in terms of FLOPs, addressing and memory access.}
% \label{tab:snn_fnn}
% \end{table}

\begin{table}[!t]
\centering
\footnotesize
\caption{ Comparison of \#FLOPs between segmentation networks on DSEC-Semantic. FR refers to firing rate of the network.}
\begin{tabular}{lccccc}
\toprule
Method & Input & \#ACC & \#MAC & Energy & FR \\ 
\midrule
ESS \cite{sun_ess_2022} (ANN) & E & 46850M &46850M &215.5 1mJ &- \\
SpikeEDN \cite{zhang_accurate_2024}(SNN) & E+F &35332M &162M &32.40mJ&  0.11 \\ \midrule
Ours (SegSNet) & F & 1586M & 60M & 1.56mJ & 0.29\\
Ours (SegMoSNet) & E & 244M & 20M & 0.31mJ & 0.35\\
Ours (Total) & E+F & 1830M & 80M & 1.87mJ & - \\
\bottomrule
\end{tabular}

\label{tab:energy_comparison_dsec}
\end{table}

\begin{table}[!t]
\centering
\footnotesize
\caption{Comparison of \#FLOPs between segmentation networks on DDD17.}
\begin{tabular}{lccccc}
\toprule
Method & Input & \#ACC & \#MAC & Energy & FR \\ 
\midrule
EV-SegSNet \cite{alonso_ev-segnet_2018}(ANN) & E & 9322M & 9322M & 42.88mJ & - \\
ESS \cite{sun_ess_2022} (ANN) & E & 11700M & 11700M & 53.82mJ & - \\
EvDistill \cite{wang_evdistill_2021} (ANN) & E & 29730M & 29730M & 136.76mJ & - \\
SpikeEDN \cite{zhang_accurate_2024}(SNN) & E+F & 7211M & 66M & 6.79mJ & 0.091 \\ \midrule
Ours (SegSNet) & F & 355M & 15M & 0.38mJ & 0.31\\
Ours (SegMoSNet) & E & 55M & 5M & 0.07mJ & 0.35\\
Ours (Total) & E+F & 410M & 20M & 0.45mJ & 0.33\\
\bottomrule
\end{tabular}

\label{tab:energy_comparison_ddd17}
\end{table}
\noindent\textbf{Energy consumption: }In addition to evaluating inter-frame performances, robustness and runtime, we also measure the power consumption of our approach, using the method followed by \cite{kim_beyond_2021}, \cite{nitin_rathi_diet-snn_2020} and \cite{zhang_accurate_2024} described in the supplementary material. The experiments include both the segmentation and motion estimation networks, on two benchmarks: DSEC-Semantic and DDD17. The results are summarized in Table \ref{tab:energy_comparison_dsec} for DSEC-Semantic and Table \ref{tab:energy_comparison_ddd17} for DDD17. Our method consistently achieves the lowest energy consumption on both datasets, reducing energy use significantly to around $1mJ$ per inference. It is important to emphasize evaluating SNN energy efficiency remains an open research topic, highly dependent on hardware. As noted by Lemaire et al.~\cite{lemaire_analytical_2023}, memory access dominates energy cost and since no prior work uses their measurements, we do not report the results in this paper.

\section{Conclusion}
In this paper, we presented a method that achieves an optimal balance between performance, runtime, energy efficiency, and inference frequency by effectively leveraging the complementary strengths of frame-based and event-based cameras. Our approach uses event streams to estimate motion between frames, eliminating the need for additional image captures and manual annotations, which are major constraints in conventional supervised training. Importantly, the method remains effective even when skipping frames, allowing accurate segmentation predictions when frames are affected by blur, overexposure, or other degradations. For future work, we aim to address the limitation of fixed event accumulation intervals, which may not adapt well to varying event densities, and to deploy our SNN-based models on neuromorphic hardware.
\section*{Acknowledgments}
\noindent This work was supported by the French Government in the Emergences Project ANR-23-PEIA-0002 managed by the National Research Agency (ANR) as part of the “PEPR IA France 2030” program.The authors are grateful to the OPAL infrastructure from Université Côte d'Azur for providing resources and support.
% \bibliography{references}

\bibliographystyle{splncs04}
\bibliography{references}
\setcounter{section}{0}
\newpage%
%\vfill
\vspace{1cm}
\begin{center}
    {\LARGE Supplementary Material}
\end{center}
\vspace{1em}

\section{Method description}
\subsection{Networks details}
\begin{figure}
    \centering
    \caption{Detailed architecture of SegSNet. "Strd" denotes the stride, H and W represent the input’s height and width and C the number of segmentation classes.}
    \includegraphics[width=0.9\linewidth]{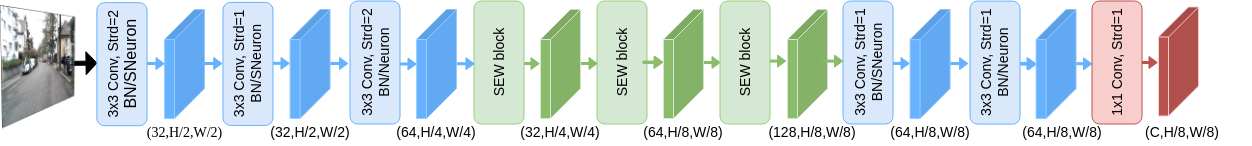}
    \label{fig:segsnet}
\end{figure}

\begin{figure}
    \centering
    \caption{Detailed architecture of SegMoSNet. The upsampling operation is performed using bilinear interpolation.}
    \includegraphics[width=0.7\linewidth]{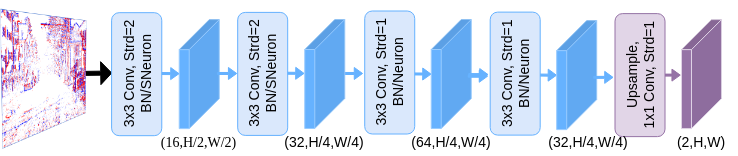}
    
    \label{fig:segmosnet}
\end{figure}

The detailed architectures of our SNN-based segmentation (SegSNet) and motion estimation (SegMoSNet) networks are shown in Fig.~\ref{fig:segsnet} and Fig.~\ref{fig:segmosnet}, respectively. The segmentation mask, named logits, assigns a class label to each pixel. It is obtained by applying the argmax operation to the output layer of SegSNet.
\vspace{-1.5cm}
\subsection{Warping function}
The warping function used in our approach to propagate the segmentation logits $S_{i,k}$ of the $k^{th}$ inter-frame segmentation, generated between consecutive frames $I_i$ and $I_{i+1}$, based on the observed motion within $\Delta t$, is defined as follows:

\begin{equation} \label{warp}
S_{i,k+1}^C(p) = \sum_q G(q, p+\delta p) S_{i,k}(q)
\end{equation}

Here,t $k$ refers to the number of the inter-frame segmentation with $k=0$ referring to the segmentation of $I_i$. $C$ indexes the number of classes and $q$ enumerates spatial positions in the logits. The variable $p$ represents a pixel location in $S_{i,k+1}$, while $p + \delta p$ corresponds to its source location in $S_{i,k}$, with $\delta p$ indicating the motion estimation of $p$ within $\Delta t$. The bilinear interpolation kernel $G(q, p+\delta p)$ weights the contribution of each neighboring pixel $q$ and is expressed as the product of two one-dimensional kernels:

\begin{equation}
G(q, p+\delta p) = g(q_x, p_x+\delta p_x) \cdot g(q_y, p_y+\delta p_y) 
\end{equation}
with $\quad g(a,b) = \max(0, 1 - |a-b|)$.

\subsection{Recurrent structure representation}
\begin{figure}[!t]
    \caption{Recurrent structure representation of the motion estimation network.}
    \centering
    \includegraphics[width=0.65\linewidth]{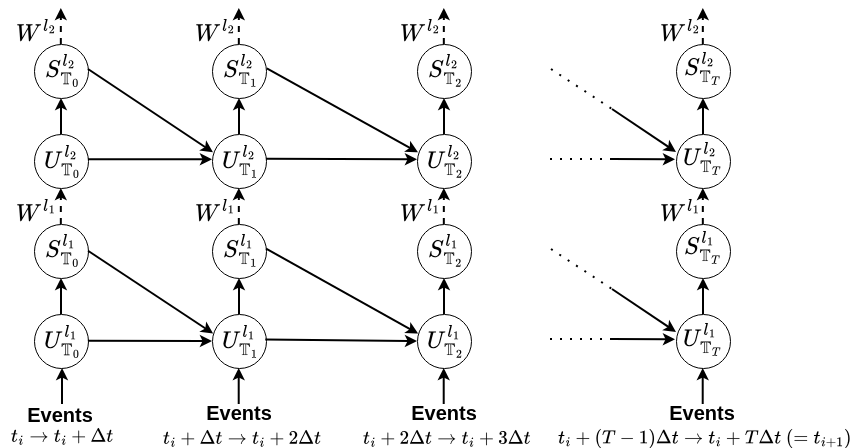}
    \label{unfold}
\end{figure}
Fig.~\ref{unfold} illustrates the recurrent structure of SegMoSNet. The membrane potential 
\( U_{\mathbb{T}_i}^{l_j} \) of a neuron in layer \( l_j \) at timestep \( \mathbb{T}_i \) is computed by combining the previous membrane potential with the spikes \( S_{\mathbb{T}_i}^{l_j} \) triggered at the previous timestep such that, if a spike occurred in that timestep, the membrane potential is reset (horizontal lines). Additionally, inputs from the previous layer are considered: when a neuron in the preceding layer fires, the corresponding synapse is activated, and its weight \( W_{l_{j-1}} \) is applied (vertical lines). Note that it shows the neurons of the first two layers.

\section{Energy measurement}
Following \cite{kim_beyond_2021}, \cite{nitin_rathi_diet-snn_2020}, \cite{zhang_accurate_2024} and \cite{li_differentiable_2021}, we compute the number of ACC operations performed by an SNN in layer $l$ using the same formulation as in these works: :  
\begin{equation}
\text{ACC}_{l} = FR_l \times T \times A_l,
\end{equation}
where $FR_l$ is the mean firing rate over the testset, defined as
\begin{equation}
FR_l = \frac{\# \text{spikes in layer } l}{\# \text{neurons in layer } l},
\end{equation}
$T$ denotes the number of timesteps, which we set to $1$ since inference is performed using a single timestep and $A_l$ is the number of addition operations in the layer $l$, given by: 
\begin{equation}
A_l = k^2 \times H_{\text{out}} \times W_{\text{out}} \times C_{\text{in}} \times C_{\text{out}}
\end{equation}
where $k$ is the kernel size, $H_{\text{out}}$ and $W_{\text{out}}$ are the output feature map dimensions and $C_{\text{in}}$ and $C_{\text{out}}$ are the numbers of input and output channels, respectively.

Although SNNs primarily rely on additions, multiplications are still required in the input layer to process integer or floating-point inputs. Therefore, the multiplication count for this layer is set equal to the corresponding $A_l$ value ($MAC_{Input}=A_l$, with $l$ is the input layer). 

For energy estimation, we adopt the widely used 45nm CMOS values \cite{horowitz_11_2014}: $0.9$pJ per ACC and $4.6$pJ per MAC and compute the energy as: 
\begin{equation}
\text{Energy}_{\text{SNN}} = \sum_l \text{ACC}_l \times 0.9 + \text{MAC} \times 4.6 
\end{equation}

% It is important to note that the current measurements consider only synaptic operations. A more comprehensive evaluation could also account for memory access, as proposed in Lemaire et al. \cite{lemaire_analytical_2023} and Damphofer et al. \cite{pimenidis_investigating_2022} which would provide a more realistic estimate of the overall computational cost.

\section{Ablation study - Why not optical flow?}
Our approach draws inspiration from video semantic segmentation \cite{zhu_deep_2017}, \cite{xu_dynamic_2018}, \cite{li_low-latency_2018}, \cite{jain_accel_2019}, where temporal consistency is achieved by propagating frame-level predictions across time. Most existing methods rely on Optical Flow (OF) to align segmentation masks between consecutive frames. However, estimating OF accurately is a challenge in practice. Supervised training \cite{gehrig_e-raft_2021} \cite{paredes-valles_taming_2023} \cite{liu_tma_2023}, \cite{gehrig_dense_2024}, \cite{wu_lightweight_2024} requires dense ground-truth, which is difficult to obtain, especially at high frame rate scenarios. In contrast, self-supervised \cite{zhu_ev-flownet_2018}, \cite{ye_unsupervised_2019}, \cite{hagenaars_self-supervised_2021}, \cite{paredes-valles_back_2021},\cite{cuadrado_optical_2023} and model-based \cite{shiba_secrets_2022} approaches, although free from supervision, generally underperform compared to supervised ones and remain computationally demanding and unsuitable for real-time deployment.
\\
To address these challenges, we replace OF with a task-specific motion estimation network, optimized directly for inter-frame segmentation rather than pixel-level displacement.  A detailed comparison between the two formulations is provided below.

% \subsection{Ablation study - Why not optical flow?}\label{ablation}
% In this section, we perform an ablation study to assess the contribution of the motion estimation component by comparing it to an optical flow. To this end, we design two sets of experiments.

\noindent\textbf{1. Comparison to optical flow ground-truth: }We train our network on $10$ms event intervals and aggregate its motion predictions over $100$ms (by summing across $10$ sub-intervals) in order to match the $100$ms ground-truth flow provided by DSEC. The comparison is performed using EPE metric (L2 endpoint error in pixels).

\begin{figure}[!t]
    \centering

    \caption{Relationship between segmentation accuracy (MIoU) and optical flow accuracy (EPE). The experiments are conducted on a subset of the training set.}
       \includegraphics[width=0.5\linewidth]{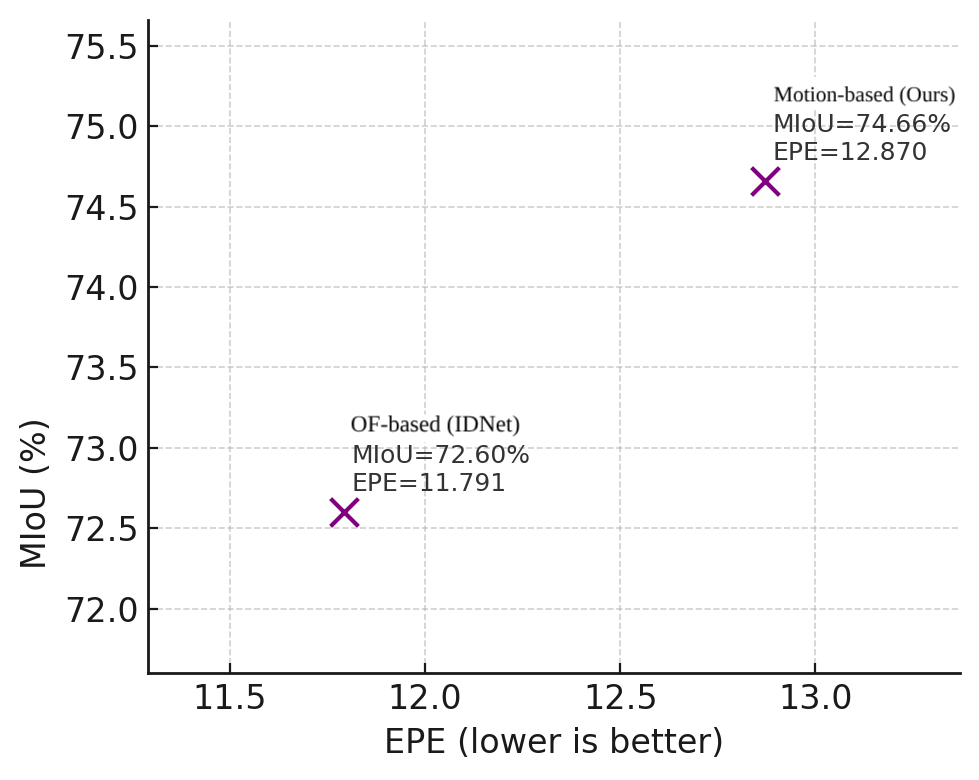}
    \label{fig:miou_vs_epe}
\end{figure}

\begin{table}[!b]
\centering
\caption{Comparison of segmentation accuracy at $100$Hz and $200$Hz, along with the average runtime per inference, between our motion estimation network (SegMoSNet) and the optical flow–based network (IDNet) \cite{wu_lightweight_2024} on the test set. Values between brackets in the MIoU column indicate the performance drop of IDNet relative to SegMoSNet, while those in the runtime column show the speed-up factor of SegMoSNet compared to IDNet.}
\begin{tabular}{l@{\hskip 6pt}c@{\hskip 6pt}c@{\hskip 6pt}cc}
\toprule
 & \multicolumn{2}{c}{MIoU} & \multicolumn{2}{c}{Runtime (ms)} \\
\cmidrule(lr){2-3} \cmidrule(lr){4-5}
 & 100Hz & 200Hz & GPU & CPU \\ 
\midrule
Motion-based (Ours)  & 56.12\%  & 55.76\% & 1.18 & 13.96 \\ 
\midrule
\multirow{2}{*}{OF-based (IDNet)}  & 55.11\%  & 54.13\% & 122 & 1350 \\ 
                                   & (-0.16) & (-1.63) & ($\times103$) & $(\times96.70)$ \\ 
\bottomrule
\end{tabular}

\label{tab:ablation_of}
\end{table}
\noindent\textbf{2. Replacing our motion estimation with OF:} To test whether conventional flow estimation could be directly integrated into our framework, we replace our motion predictor with IDNet\cite{wu_lightweight_2024}, a recent state-of-the-art optical flow network achieving the best trade-off between runtime and performances by using an ANN with $1.4$M parameters which predicts $100$ms flow estimation. 
\\
These experiments are designed to clarify two key points: (i) whether our learned motion predictions correspond to conventional optical flow and (ii) whether using an optical flow network improves segmentation propagation in our framework. The corresponding results are presented in Fig.\ref{fig:miou_vs_epe}. This figure shows a comparison between our motion estimation network with IDNet in terms of segmentation accuracy (MIoU) and flow accuracy (EPE). For a fair comparison, IDNet produces $10$ms predictions, which are accumulated over $100$ms before being compared with the ground-truth flow using EPE, mirroring our setup. MIoU is computed from motion or flow predictions every $10$ms, evaluated against segmentation ground truth every two frames (instead of every frame, as in our previous experiments). This evaluation is conducted on a subset of the DSEC training folds where both segmentation and flow ground truth are available (\textit{zurich$\_$city$\_$01$\_$a}, \textit{02$\_$a}, \textit{05$\_$a}, \textit{06$\_$a}, \textit{07$\_$a} and \textit{08$\_$a}). Further MIoU results on the test set, along with the runtime of each model, are provided in Table~\ref{tab:ablation_of}.

\noindent The figure shows that SegMoSNet (ours) achieves a worse (higher) EPE ($12.87$) while achieving a strong segmentation performance with MIoU $74.66\%$ and OF-based (IDNet) yields a better (lower) EPE ($11.79$) but also a lower MIoU. This result highlights that our motion estimation model provides more reliable performance within our framework compared to the optical flow-based alternative, while requiring $35\times$ fewer parameters.
\\
In addition, the results summarized in Table~\ref{tab:ablation_of} are consistent with ~\ref{fig:miou_vs_epe}. They show that SegMoSNet achieves higher MIoU than the optical flow–based approach at both $100$Hz (predictions every $10$ms) and $200$Hz (predictions every $5$ms). While being significantly more efficient and running over two orders of magnitude faster than IDNet on both GPU and CPU.

\end{document}